\documentclass[10pt,twocolumn,letterpaper]{article}

\usepackage{3dv}
\usepackage{times}
\usepackage{epsfig}
\usepackage{graphicx}
\usepackage{amsmath}
\usepackage{amssymb}
\usepackage{tabularx}
\usepackage{lipsum}
\usepackage{booktabs}

\usepackage[breaklinks=true,letterpaper=true,colorlinks,bookmarks=false]{hyperref}

\threedvfinalcopy 

\def\threedvPaperID{5} 

\ifthreedvfinal\fi
\begin{document}

\title{Robust 3D Human Motion Reconstruction Via Dynamic Template Construction}

\author{Zhong Li$^{\text{1,2}}$ \qquad Yu Ji$^{\text{2}}$ \qquad  Wei Yang$^{\text{1,2}}$ \qquad  Jinwei Ye$^{\text{3}}$ \qquad Jingyi Yu$^{\text{1,2,4}}$ \\
\and
$^{\text{1}}$University of Delaware,Newark,USA  \\
{\tt \small lizhong@udel.edu}
\and
$^{\text{2}}$Plex-VR  \\
{\tt \small \{yu.ji,wei.yang\}@plex-vr.com}
\and
$^{\text{3}}$Louisiana State University,Baton Rouge,USA  \\
{\tt \small jinweiye@lsu.edu}
\and
$^{\text{4}}$ShanghaiTech University,Shanghai,China  \\
{\tt \small yujy1g@shanghaitech.edu}
}

\maketitle

\begin{abstract}
  In multi-view human body capture systems, the recovered 3D geometry or even the acquired imagery data can be heavily corrupted due to occlusions, noise, limited field-of-view, etc. Direct estimation of 3D pose, body shape or motion on these low-quality data has been traditionally challenging.In this paper, we present a graph-based non-rigid shape registration framework that can simultaneously recover 3D human body geometry and estimate pose/motion at high fidelity.Our approach first generates a global full-body template by registering all poses in the acquired motion sequence.We then construct a deformable graph by utilizing the rigid components in the global template.We directly warp the global template graph back to each motion frame in order to fill in missing geometry. Specifically,we combine local rigidity and temporal coherence constraints to maintain geometry and motion consistencies.Comprehensive experiments on various scenes show that our method is accurate and robust even in the presence of drastic motions.

\end{abstract}

\vspace{-10px}
\section{Introduction}
\label{sec:1}
\vspace{0px}
Despite tremendous efforts and advances in 3D shape and motion reconstruction~\cite{debevec2012light,starck2003model,vlasic2009dynamic,starck2007surface,orts2016holoportation,budd2013global,zhang2014quality,wang2016capturing},
reliable estimation of 3D pose, body geometry and motion trajectory remains challenging. 
3D reconstruction produced by traditional photogrammetry or multi-view geometry can be heavily corrupted due to occlusions, noise, limited field-of-view, etc.
It is possible to add additional cameras to improve the reconstruction but would lead to higher computational and equipment cost. One possible solution is to complete the
missing data via geometric operators such as filtering and hole filling (\eg, Poisson
 surface completion)~\cite{kazhdan2013screened,kazhdan2006poisson,liepa2003filling}.
By far these methods can only handle small holes. It is also possible to adopt a template-based approach~\cite{li2009robust,zollhofer2014real,guo2017robust} by using
pre-reconstructed full body 3D geometry. In reality, generating the template requires special acquisition system that is inaccessible to commodity users. Further, such techniques
cannot handle 
strong deformations caused by drastic motion.


\begin{figure*}
  \centering
  \includegraphics[scale=0.25]{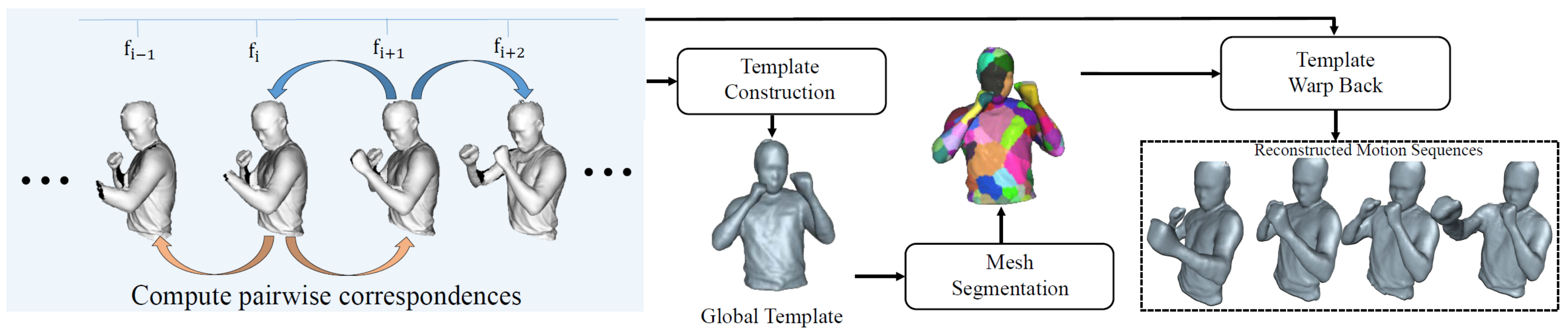}
  \caption{A diagram showing the pipeline of our reconstruction algorithm.}\label{fig:pipeline}
\end{figure*}

In this paper, we present a graph-based non-rigid shape registration framework that can simultaneously recover
3D human body geometry and estimate motion at high fidelity. Our approach first generates a global full-body
template by registering all poses in the acquired motion sequence. We observe that missing body geometry in one frame may
appear in other frames in the motion sequence. This implies that we can generate the complete body template by aligning each individual
partial reconstruction. To do so, we conduct multi-frame correspondence matching by imposing a temporal coherence constraint.
We consider both forward and backward motions to formulate the temporal regularization. We then construct a deformable graph by utilizing the rigid components in the global template. Although the human body is non-rigidity, it can be effectively decomposed into piece-wise rigid components. We hence segment the global template into connected rigid patches and build a deformable graph with centroid of rigid patches as nodes. Finally, we develop a patch surface expansion approach for fitting the global template in terms of each node's motion estimation. We also impose temporal consistency to maintain local rigidity and motion smoothness. The reconstruction pipeline of our algorithm is shown in Fig.~\ref{fig:pipeline}.
Comprehensive experiments on a multi-view system show that our method is accurate and robust even in the presence of drastic motions.
\vspace{0px}
\section{Related Work}
\label{sec:2}
\vspace{0px}
There is an emerging trend on using multi-view acquisition techniques for reconstructing 3D human body geometry and motion.
Notable examples include techniques using a multi-view camera system~\cite{orts2016holoportation,bogo2015detailed,mustafa2016temporally,mustafa2015general},shape-from-silhouette~\cite{gkalelis2009i3dpost,cheung2003shape}, multiple-view stereo matching~\cite{schonberger2016pixelwise}, and photometric stereo~\cite{vlasic2009dynamic,debevec2012light}. The focus has been on conducting non-rigid registration~\cite{huang2008non,li2008global} for mesh sequence tracking and 3D reconstruction from the captured data. Most previous work falls into two categories, \ie template-free shape alignment approach and template-based deformation approach.

\textbf{Template-free shape alignment.} This class of methods performs per-frame reconstruction without using a global full body template. S\"{u}ssmuth  \etal \cite{sussmuth2008reconstructing} map all input scans onto a 4D space-time volume and conduct high-dimensional shape reconstruction. Mitra \etal \cite{mitra2007dynamic} also use a 4D space-time representation to compute the motion of the scanned object. They recover the volume by estimating globally consistent motion instead of pairwise alignment. Wand \etal \cite{wand2007reconstruction} applied a statistic framework to conduct pairwise shape alignment if the topology remains consistent. They further improves the template-free shape alignment by using volumetric deformation model \cite{wand2009efficient}. However, the computational complexity is very high and the performance is limited by the running time. The algorithm is also sensitive to corrupted input data such as large shape deformation and/or truncated geometry. To summarize, above template-free approaches can only handle small motion due to the accumulation of tracking errors.

\textbf{Template-based approaches.} This class of methods attempts to utilize geometric template as a shape prior for mesh sequences tracking. Some focus on tracking and reconstructing the model to accommodate general scenarios. Offline approaches such as \cite{li2009robust} acquire a coarse low-resolution template via static acquisition and then track the input sequence using embedded deformation \cite{sumner2007embedded}. Dou \etal \cite{dou2013scanning} use an eight-depth camera system to reconstruct the full body geometry and track the motion by deforming a pre-captured human body template. Zollh\"{o}fer \etal \cite{zollhofer2014real} perform online template acquisition for mesh tracking and use GPU acceleration to achieve real-time performance. However, acquiring the online template requires the motion to be rigid and is prone to errors in case of drastic motions. Newcombe \etal \cite{newcombe2015dynamicfusion} extended the Kinect fusion algorithm \cite{newcombe2011kinectfusion} to perform template-based reconstruction. Their approach is able to capture the non-rigid partial views of a moving person. However, their system can only handle relatively slow motion. \textcolor[rgb]{0.00,0.00,0.00}{Guo \etal~\cite{guo2017robust} use L0 based regularizer to achieve more accurate and robust result.} \textcolor[rgb]{0.00,0.00,0.00}{More recent approaches \cite{collet2015high,prada2016motion,huang2015hybrid,budd2013global} adopt a keyframe-based mesh tracking and similarity tree scheme and are able to handle topology changes and significantly reduce the tracking failure rate.}

Other non-rigid tracking approaches tackle elastic objects and are suitable for emulating Cartoon style avatars. Vlasic \etal \cite{vlasic2008articulated} apply shape-from-silhouette and deformed a statically acquisition template via linearly blended skinning \cite{lewis2000pose}. Huang \etal\cite{huang2013robust} use a skeleton-based hybrid deformation approach. \textcolor[rgb]{0.00,0.00,0.00}{Rhodin \etal \cite{rhodin2016general} and Robertini \etal\cite{robertini2016model} present pleasant results on outdoor motion capture, however, their methods are based on articulated skeleton thus can't applied to general shape.} Cargniart \etal \cite{cagniart2010free} propose a patch-based approach.~\cite{yang2016computing} and ~\cite{bogo2015detailed} explore fitting 3D body model database onto the acquired data. Similar methods have been also applied to face and hand tracking~\cite{li2013realtime,cao20133d,qian2014realtime}. Another seminal work of Holoportation by Dou \etal \cite{dou2016fusion4d} achieves real-time performance capture,however, their results are sensitive to background segmentation errors.

Our approach falls into the category of template-based approach. However, different from ~\cite{zollhofer2014real,li2009robust}, we do not require a separate process for building the template. Instead, we construct our global template by accumulating individual frames during the capture process. Our system uses a multi-view stereo capture system for data acquisition. However, our input data is corrupted due to viewing frustum truncation and drastic motion. Direct reconstruction from multi-view stereo approach exhibits large holes and even truncations. In our approach, we propose to exploit the temporal redundancies to solve this problem.


\section{3D Human Shape/Motion Reconstruction}
\label{sec:3}
 \vspace{-5px}
Our algorithm consists of four major steps. We first build a global human body template from a motion sequence with incomplete body geometry. In order to achieve this, we establish pairwise correspondences between adjacent motion frames by imposing a temporal regularization term. By minimizing our global deformation energy function, we align the incomplete poses from all frame to a global template. Next, we use the graph-cut algorithm to segment the global template into multiple connected rigid patches and use the segmentation results to determine the global nodes. Finally, we estimate the rotation parameters to warp the piece-wise rigid global template back to each input frame in order to recover the full body geometry for the entire motion sequence.
\subsection{Pairwise Surface Matching}
\label{sec:3.1}
To build the global template, we first need to register the surfaces from adjacent motion frames.
We use a deformation graph technique similar to \cite{sumner2007embedded}.
Given a sequence of captured $N$ motion frames
$\{\boldsymbol{P}^n|_{n =1,...,N}\}$, where a frame $P^n$ has $\kappa$ vertices $\{v_i |_{i = 1,...,\kappa}\}$, where $v_i\in\mathbb{R}^3$,
 we first uniformly sample a set of graph nodes $\boldsymbol{G} = \small\{g_1,g_2,...,g_m\small\}$ (where $m << \kappa$) on the surface $P^n$.  Once we have graph nodes, we use the deformation of graph nodes to represent the movement of vertex. Specifically, we use affine transformation $\boldsymbol{A} = \small\{\boldsymbol{A}_t\small\}_{t=1}^m$ and $\boldsymbol{b} = \small\{\boldsymbol{b}_t\small\}_{t=1}^m$ to parameterize the deformable movement of a graph node. After deformation, the new position of a vertex $v$ can be written as:
 \vspace{-5px}
	\begin{equation}
    \label{eq:affine}
    v' = f(v,\boldsymbol{A},\boldsymbol{b}) = \sum_{t=1}^m w_t(v)[\boldsymbol{A}_t(v-g_t)+g_t+\boldsymbol{b}_t]
    \end{equation}

where $w_t(v) $ is the weighing factor of a graph node $g_t$ on the vertex $v$. In particular, $w_t (v) = max(0,(1-d^2(v,g_t)/r^2)^3)$, where $d(v,g_t)$ is geodesic distance between $v$ and $g_t$ and $r$ is the distance between $v$ and its K-nearest nodes in the geodesic distance domain (we use K=4 in our experiments).

Once we have constructed the deformation graph, we align the surface onto other frames by finding the optimal affine transformation of its graph nodes. Recall that our input is a sequence of deformable surfaces. To align all the surfaces, a brute-force approach is to use non-rigid registration~\cite{chui2003new}. A major drawback of using this approach is the lack of stability: deformation errors would accumulate over the frames and can result in failure of the algorithm.

\begin{figure}
\centering
\includegraphics[width=.65\linewidth]{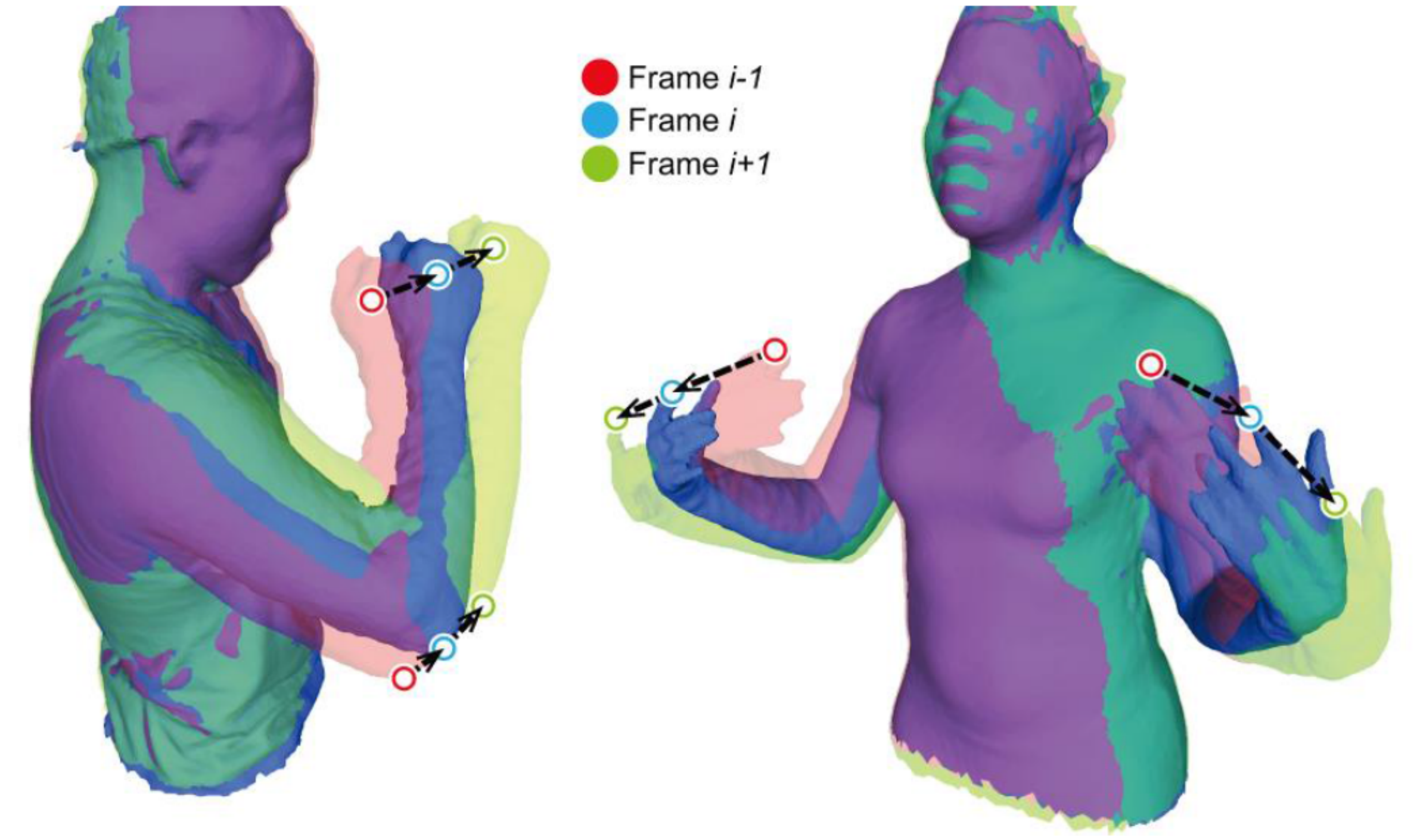}
\caption{Human motions exhibit temporal smoothness between adjacent frames.}
\label{fig:temporal_hint} 
\end{figure}

An alternative solution is to perform pairwise correspondence matching ~\cite{li2009robust}. This approach attempts to find correspondences between adjacent meshes. Compared with their inputs, our surface sequences are more challenging due to incomplete geometry and drastic motions.
 As shown in Fig. \ref{fig:temporal_hint}, although largely overlapped, adjacent surface meshes exhibit temporal smoothness between adjacent motions. Furthermore, the non-rigidity of human body geometry can cause large errors even in the presence of small motions since affine transform is no longer sufficient to characterize the motion. We propose to solve these challenges by exploiting the temporal coherence. As shown Fig.~\ref{fig:pipeline}, adjacent motion frames are highly consistent due to the motion smoothness. We therefore add a temporal smoothness term to the pairwise correspondence energy function in order to enforce the motion continuity. In particular, we register three consecutive frames (\ie we consider both forward and backward motion) at the same time. As shown in Fig.~\ref{fig:pipeline}, we warp a frame $i$ onto its previous ($i-1$) and successive ($i+1$) frames. Therefore, our pairwise correspondence matching energy function is defined as:

    \begin{equation}
    \label{eq:total}
    \boldsymbol{E}_{total}=\lambda_1\boldsymbol{E}_{rigid}^{\pm } + \lambda_2\boldsymbol{E}_{smooth}^{\pm}
    \\ + \lambda_3\boldsymbol{E}_{fit}^{\pm}+ \lambda_4\boldsymbol{E}_{tempo}
    \end{equation}
In this equation, we omit the frame stamp $n$ in superscript and use "$+$" for forward motion "$n \rightarrow n+1$" and "$-$" for backward motion "$n \rightarrow n-1$". $\lambda_1  \sim \lambda_4$ are weighing factors for balancing the regularization terms. In our experiments, we use $\lambda_{1} = 100$, $\lambda_{2} = 20$, $\lambda_{3} = 1$ and $\lambda_{4} = 5$. Next, we explain each energy term in Eqn.~\ref{eq:total} in details.

The first term $\boldsymbol{E}_{rigid}$ constraints the rigidity enforced by the affine transformation,and thus is defined as:
      \begin{multline}
       \label{eq:rigid}
        \boldsymbol{E}_{rigid}= \sum_{G} \big((\boldsymbol{a}_1^{T}\boldsymbol{a}_2)^2+(\boldsymbol{a}_2^{T}\boldsymbol{a}_3)^2+(\boldsymbol{a}_1^{T} \boldsymbol{a}_3)^2\\
            					+(1-\boldsymbol{a}_1^{T}\boldsymbol{a}_1)^2+(1-\boldsymbol{a}_2^{T}\boldsymbol{a}_2)^2+(1-\boldsymbol{a}_3^{T} \boldsymbol{a}_3)^2\big)
             \end{multline}
where $\boldsymbol{a}_1$, $\boldsymbol{a}_2$ and $\boldsymbol{a}_3$ are the three column vectors that form the $3\times3$  matrix $\boldsymbol{A}_t$.

\textcolor[rgb]{0.00,0.00,0.00}{
The second term $\boldsymbol{E}_{smooth}$ enforces the spatial smoothness of the geometric deformation in one frame and it is written as:}

		\begin{equation}
        \label{eq:smooth}
		\begin{split}
			\boldsymbol{E}_{smooth} = \sum_{t=1}^m\sum_{k\in\nu(t)}\hat{w}_{(t,k)}||\boldsymbol{A}_t(g_k-g_t)+g_t+\\
            \boldsymbol{b}_t-(g_k+\boldsymbol{b}_k)||^2
        \end{split}
        \end{equation}
where $\nu(t)$ is node $g_t$'s neighbor that shares the same edge in the sub-sample graph.

We adopt a data fitting term ${E}_{fit} $ similar to Iterated Closest Point (ICP)  to measure vertex displacements between the reference frame and the target frame. The fitting cost consists of two components: one for minimizing the point-to-point distances and the other for minimizing the point-to-plane distances. Further, instead of using the closest points as correspondences, we trace an undirected ray $n_i$ along the normal direction of the source vertex $v_i$ and choose the vertex that is the closest to the ray-target surface intersection as the temporary correspondence $c_i$:
	\begin{equation}
	   \boldsymbol{E}_{fit} = \sum_{i\in P}\lambda_{point}||v_i-c_i||^2 + \lambda_{plane}|n_i^T(v_i-c_i)|^2
    \end{equation}
In our experiments, we use $\lambda_{point}=0.1$ and $\lambda_{plane}=1$.

Finally, we propose a temporal regularization term $\boldsymbol{E}_{tempo}$ to preserve the motion continuity among three consecutive frames, \textit{i.e.} from frame $n$ to frame $n-1$ as well as frame $n+1$. More specifically, we constrain the current-to-next motion $\small\{{A}^{+}_t,\boldsymbol{b}^{+}_t\small\}_{t=1}^m$ by current-to-previous motion $\small\{{A}^{-}_t,\boldsymbol{b}^{-}_t\small\}_{t=1}^m$. Since motions between adjacent frames are similar, we formulate a new energy term $\boldsymbol{E}_{tempo}$ to force ${A}^{-}_t{A}^{+}_t$ close to an identity matrix, and minimize ${A}^{-}_t\boldsymbol{b}^{+}_t+\boldsymbol{b}^{-}_t$:

	\begin{equation}
    \label{eq:tempo}
    \boldsymbol{E}_{tempo} = \sum_{t=1}^m ||\mathbb{I} - A_t^+A_t^-||_F^2
                                +||A_t^-b_t^+ + b_t^-||_2^2
    \end{equation}
where $\mathbb{I}$ is an identity matrix.

In our implementation, we solve Equation.~\ref{eq:total} in an iterative manner by using the Gauss-Newton method.

\begin{figure}
  \centering
  \includegraphics[width=.9\linewidth]{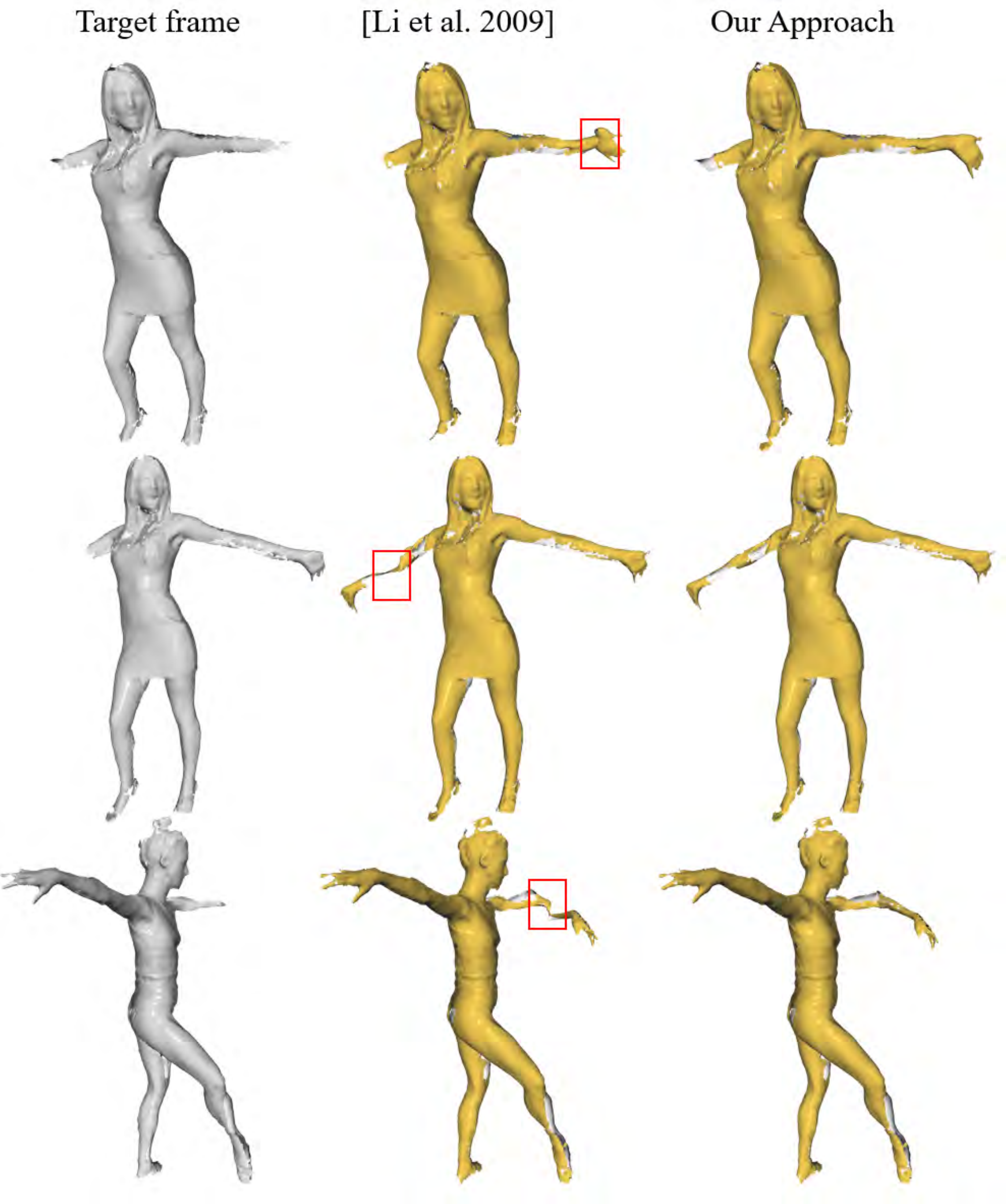}
  \caption{Pairwise correspondence matching results. We show the target frames in the first column. We compare our registration results (third column) with results of \cite{li2009robust}  (second column). }
  \label{fig:pairwise_compare}
\end{figure}


%

To illustrate the effectiveness of pairwise correspondence optimization algorithm, we show our frame alignment results in Fig.~\ref{fig:pairwise_compare} and compare with \cite{li2009robust}. Notice that the input frames exhibit severe occlusions and/or geometric truncations. Our algorithm still generates accurate alignment results with fewer artifacts due to the consideration of temporal smoothness term.

\subsection{Global Template Construction}
\label{sec:3.2}
Recall that our input frames are incomplete and exhibit many missing parts due to occlusions/truncations. We observe that the occluded geometry may appear in later frames as the pose changes. So we set out to align all input frames $\{\boldsymbol{P}^n|_{n =1,...,N}\}$ into an optimized pose $\boldsymbol{P}^0$ where nearly all occluded regions are filled. Notice that we have already obtained dense correspondences using the optimization algorithm described in Section~\ref{sec:3.1}, similar to \cite{li20133d,dou20153d},we further define an energy function $E_{global}$ as follow to construct a global template:
\begin{equation}
    \label{eq:global}
	   \boldsymbol{E}_{global} = \sum_{n=1}^N(\lambda_{r}\boldsymbol{E}_{rigid}^{n  \rightarrow 0} + \lambda_{s}\boldsymbol{E}_{smooth}^{n  \rightarrow 0})	+ \lambda_{c}\boldsymbol{E}_{corr}
\end{equation}
where $\boldsymbol{E}_{rigid}$ and $\boldsymbol{E}_{smooth}$ are the same as in Eqn.~\ref{eq:total}. $\boldsymbol{E}_{corr}$ is a data term to impose the distant consistency between corresponding vertices in adjacent frames. $\boldsymbol{E}_{corr}$ is defined as:
\begin{equation}
    \label{eq:corr}
    \begin{split}
	   \boldsymbol{E}_{corr} = \sum_{n=1}^{N-1}||f(\boldsymbol{P}^n,\boldsymbol{A}^{n  \rightarrow 0},\boldsymbol{b}^{n  \rightarrow 0})\\
        -f(f(\boldsymbol{P}^n,\boldsymbol{A}^{+},\boldsymbol{b}^{+}),\boldsymbol{A}^{n+1  \rightarrow 0},\boldsymbol{b}^{n+1  \rightarrow 0})||^2
    \end{split}
\end{equation}
where $f(\cdot)$ is the deformed position from Eq.\ref{eq:affine}. 

In our experiment, we use $\lambda_{r}=150$, $\lambda_{s}= 5$ and $\lambda_{c}=1$. We iteratively solve the equation via  Gauss-Newton optimization to sequentially align consecutive frames to obtain a global optimal alignment.

Once we align all input frames, we then "stitch" them together to form the final global template. Notice that directly fusing the point clouds can lead to large errors such as discontinuity. We instead fuse their gradients and then reintegrate the surface. The process is analogous to image completion in the gradient domain and in our solution we apply poisson surface reconstruction~\cite{kazhdan2006poisson} to obtain the reconstructed template mesh.
\begin{figure}
\centering
\includegraphics[width=.9\linewidth]{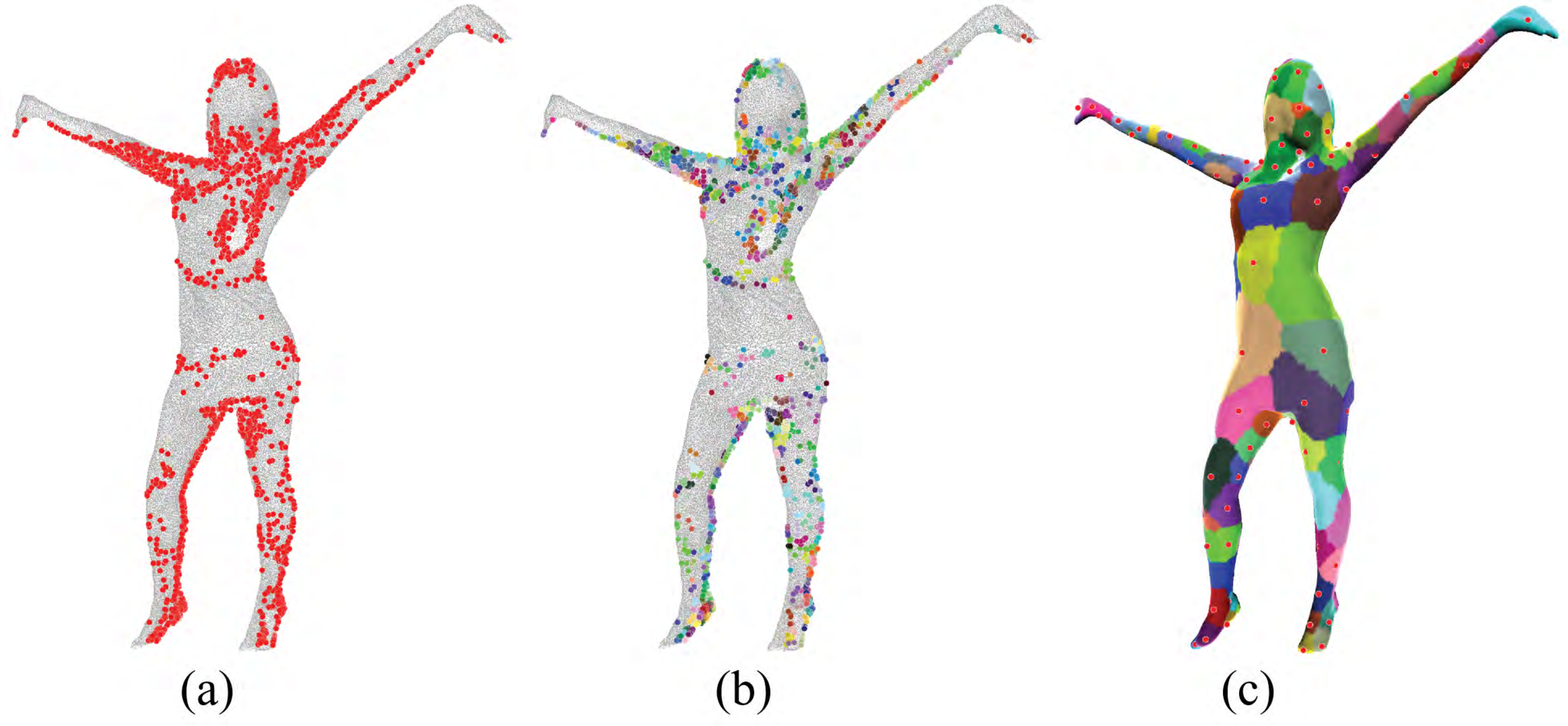}
\caption{Patch segmentation. (a) All aligned graph nodes in the global template; (b) Grouped nodes (color-coded) after K-means clustering; (c) Final segmentation result after graph-cut. }
\label{fig:patch_illustration}
\end{figure}

\subsection{Patch Segmentation}
\label{sec:3.3}
Once we have the global template, we map all input frames $\{\boldsymbol{P}^n|_{n =1,...,N}\}$ onto the global template mesh $\boldsymbol{P}^0$ through a common deform graph $\boldsymbol{G}^0$. We assume that the topology (e.g., the number of nodes and edge connectivity) remain consistent across frames. Specifically, we segment the global template mesh $\boldsymbol{P}^0$ into patches and treat the geometry of each patch relatively rigid. We then use the centroid of each patch as the node in the global deform graph $\boldsymbol{G}^0$. In contrary to ~\cite{cagniart2010free} in which the patch segmentation is performed based on geodesic distance, we also consider the motion similarity among vertices.

In particular, we set out to partition the vertices $\widetilde{v}$ in $\boldsymbol{P}^0$ into relatively rigid subsets. For an input frame $\boldsymbol{P}^n$, we use $v^\prime$ and $g^\prime$ to represent the vertex and graph node respectively after the global registration. We then perform K-means clustering for all aligned graph nodes according to their Euclidean distances. We set the pre-defined number of clusters $\boldsymbol{K}$ as the maximum number of deform graph nodes in all $N$ frames.

For each vertex $\widetilde{v_i}$ in global template mesh $\boldsymbol{P}^0$, we first find its K-nearest neighbors $\Omega(\widetilde{v_i})$ in the aligned vertices $v^\prime$ of all frames $\{\boldsymbol{P}^n|_{n =1,...,N}\}$. We then calculate the weight between $\widetilde{v_i}$ and cluster $c_j$ using the mean value of all weights between vertices $v^\prime$ in $\Omega(\widetilde{v_i})$ and graph nodes $g^\prime$ in $c_j$:

    \begin{equation}
    \begin{split}
   w(\widetilde{v_i}, c_j) = \sum_{v^\prime \in \Omega(\widetilde{v_i})} \sum_{g^\prime \in c_j} w(v^\prime, g^\prime)/S
    \end{split}
    \end{equation}
where $S$ is the total number of valid $w(v^\prime, g^\prime)$.

Since $w(v^\prime, g^\prime)$ corresponds to the weighing factor of a graph node $g^\prime$ on vertex $v^\prime$, we can also use $w(\widetilde{v_i}, c_j)$ to determine how significance of cluster $c_j$ with respect to $\widetilde{v_i}$. The set of vertices affected most by the same cluster should have a relative similar rigid motion. Therefore, we can simply treat weight $w(\widetilde{v_i}, c_j)$ as the data cost for assigning $\widetilde{v_i}$ to cluster $c_j$. We further use the pots form smoothness cost: $p(\widetilde{v_i}, \widetilde{v_k}) = 0$ when $\widetilde{v_i}, \widetilde{v_k}$ have the same label and belong to the same triangle in $\boldsymbol{P}^0$ and 1 otherwise. Finally, we formulate the energy function as:
    \begin{equation}
    \begin{split}
        E = -\sum_{\widetilde{v_i}} w(\widetilde{v_i}, c_j) + \lambda \sum_{\{\widetilde{v_i}, \widetilde{v_k}\} \in  \mathcal{N}}p(\widetilde{v_i}, \widetilde{v_k})
    \end{split}
    \end{equation}
where $\lambda$ is a weighting factor. To find an optimal solution, we apply the graph-cut algorithm \cite{delong2012fast} and we group vertices with the same label into a patch. An example of segmentation result is shown in Fig.~\ref{fig:patch_illustration}.

\begin{figure}
\centering
\includegraphics[width=\linewidth]{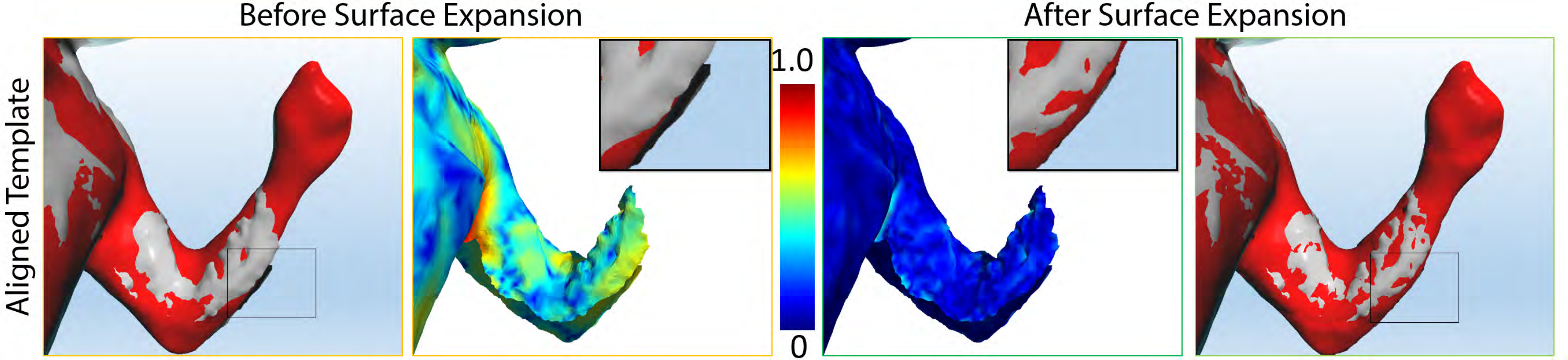}
\caption{Surface expansion result. Our patch expansion algorithm can effectively reduce the misalignment between deformed frames and the global template.}

\label{fig:surface_expansion}
\end{figure}

\begin{figure}
\centering
\includegraphics[width=.8\linewidth]{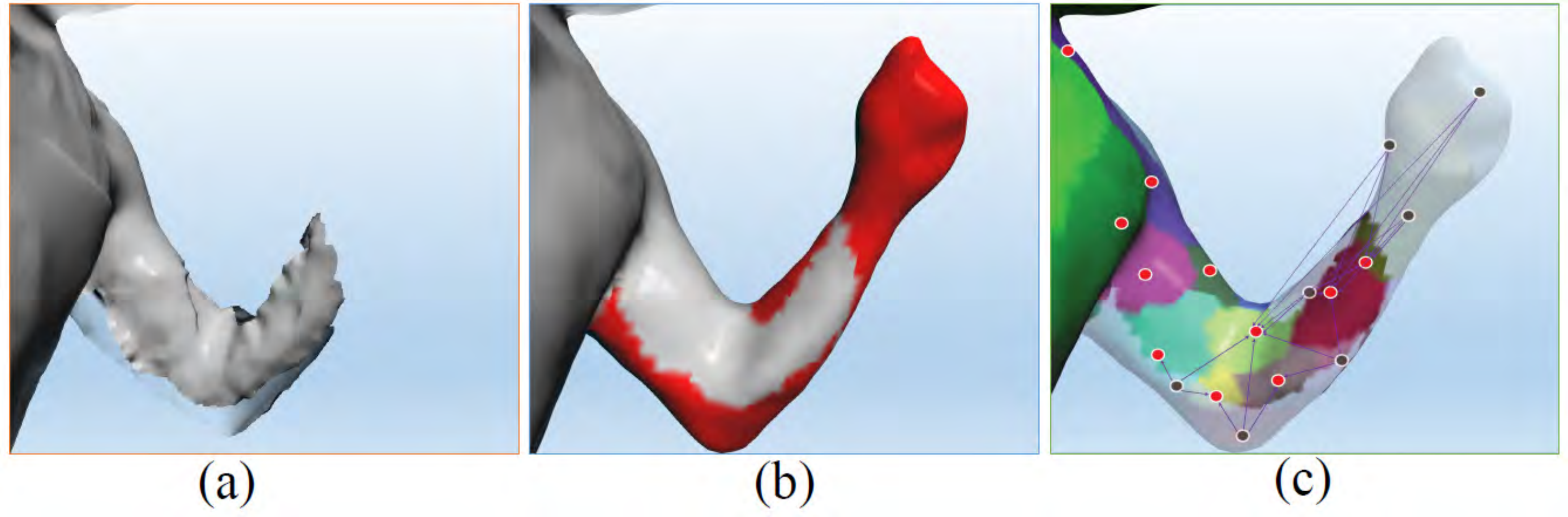}
\caption{Weighted node estimation using geodesic distance. (a) original input mesh which is truncated; (b) the global template after
surface expansion, Where corrupted regions are marked in red; (c) our weighted node estimation, where each black node is weighed by its distances between three closet neighbors (red nodes).}
\label{fig:weighted_intial_approximation}
\end{figure}
\vspace{0px}
\subsection{Surface Expansion and Patch Warping}
\label{sec:3.4}
	Once we partition the global template $\boldsymbol{P}^0$ into $\boldsymbol{K}$ patches, We treat each patch $patch_\tau$'s centroid $g_t$ as the graph node.

\vspace{0px}

To warp the global template $\boldsymbol{P}^0$ back to each frame $\boldsymbol{P}^{n}$, we adopt a two-step approach to first approximate and then optimize the graph motion parameter $\small\{\small\{A_t^{0,n},b_t^{0,n}\small\}_{t=1}^{\boldsymbol{K}}\small\}_n^N$ respectively. We first conduct the closest point approximation, second step is to further optimize them which is constrain by adjacent temporal information.

\vspace{0px}

\par Recall that we have already aligned each input frame $\small\{\boldsymbol{P}^{n}\small\}_{n=1}^N$ to an optimal position $\small\{\boldsymbol{P}^{n \rightarrow 0}\small\}_{n=1}^N$ when building the global template. We can thus directly convert $\boldsymbol{P}^{n \rightarrow 0}$'s graph node's motion $\small\{\boldsymbol{A}_t,\boldsymbol{b}_t\small\}_{t=1}^m$ to each vertex $v_i$'s rigid rotation $\boldsymbol{R}_i$ and translation $\boldsymbol{T}_i$ by further decompose Eq. \ref{eq:affine}.




Every vertex $v_i^n$ in $\boldsymbol{P}^{n}$ can be viewed to go through a rigid motion $\boldsymbol{R}_i^{n \rightarrow 0},\boldsymbol{T}_i^{n \rightarrow 0}$ to an optimized target $v_i^{n \rightarrow 0}$ after deformation and we can then warp $v_i^{n \rightarrow 0}$ back through $v_i^n = \boldsymbol{R}_i^{0 \rightarrow n}v_i^{n \rightarrow 0} + \boldsymbol{T}_i^{0 \rightarrow n}$, where $\boldsymbol{R}_i^{0 \rightarrow n} = inv(\boldsymbol{R}_i^{n \rightarrow 0})$ and $\boldsymbol{T}_i^{0 \rightarrow n} = -inv(\boldsymbol{R}_i^{n \rightarrow 0})\boldsymbol{T}_i^{n \rightarrow 0}$. To approximate the each $g_t$'s motion parameter when warping it to $\boldsymbol{P}^{n}$, we locate the graph node $g_t$'s closest point $v_{i^{\prime}}^{n \rightarrow 0}$ in $\boldsymbol{P}^{n \rightarrow 0}$ and use $v_{i^{\prime}}^{n \rightarrow 0}$'s $\small\{\boldsymbol{R}_i^{0 \rightarrow n},\boldsymbol{T}_i^{0 \rightarrow n}\small\}$ as the motion parameter. However, from Fig.~\ref{fig:surface_expansion}, we observe that the deformed frame and the reconstructed global template can still exhibit relatively large misalignments. To better approximate the motion parameters, we present a patch based surface expansion approach based on \cite{zhang2008spacetime} to better fit global template onto the deformed frame $\boldsymbol{P}^{n \rightarrow 0}$:
	\begin{equation}
	\begin{split}
	\boldsymbol{E}_{expan} = \sum_{\gamma\in T_k} ||v_\gamma^0 + d_\gamma n_\gamma^0 - c_\gamma||^2  \\
                    + \lambda_{patch}\sum_{patch_\tau}^{\boldsymbol{K}}\sum_{\nu \in patch_\tau} \sum_{k \in \eta(\nu_\gamma)} |d_i - d_k|^2
    \end{split}
    \end{equation}

Specifically, we trace a ray from each vertex $v_\gamma^{0}$ on the global template mesh along its normal direction $n_\gamma^{0}$ to the target deformed mesh $\boldsymbol{P}^{n \rightarrow 0}$. Since $\boldsymbol{P}^{n \rightarrow 0}$ may be truncated due to occlusion, not all $v_i^0$ will be able to find intersections with the deformed mesh $\boldsymbol{P}^{n \rightarrow 0}$. We denote the ones we manage to find the corresponding points as $\small\{v_\gamma^0, c_\gamma, n_\gamma^0\small\}_{\gamma \in T_k}$. The first term of the $\boldsymbol{E}_{expan}$ aims to minimize the distance $d_\gamma$ between vertex $v_\gamma^0$ and its intersection point $c_\gamma$. The second part regularization term ensures smoothness. We enforce it by setting $d_i$ close to its K-nearest neighbors $d_k$ in its patch $patch_\tau$. It will also propagate $d$ to non-correspondence vertex. Fig.~\ref{fig:surface_expansion} shows the results before and after optimizing $d_i$.

A further comparison between the expanded global template $\boldsymbol{P}_{expan}^n$ with each deformed input frame $\boldsymbol{P}^n$ \textcolor[rgb]{0.00,0.00,0.00}{as shown in Fig.~\ref{fig:surface_expansion}} illustrates that the expanded global template recovers the occluded parts. We again trace a ray from the global node $g_t$ along its normal direction to determine whether $g_t$ intersects with the target deformed frame $\boldsymbol{P}^{n \rightarrow 0}$. If yes, we adopt $\small\{\boldsymbol{R}_t^{0 \rightarrow n} , \boldsymbol{T}_t^{0 \rightarrow n}\small\}$ and further convert it to $\small\{\boldsymbol{A}_t^{0 \rightarrow n},\boldsymbol{b}_t^{0 \rightarrow n}\small\}$ from its closest point in $\boldsymbol{P}^{n \rightarrow 0}$ as nodes motion parameter. If not, we approximate its motion parameter as weighted average of its K-nearest (K=3 in our experiment) known motion parameters where the weights correspond to geodesic distance, as shown in Fig.~\ref{fig:weighted_intial_approximation}.



\begin{figure}
\centering
\includegraphics[width=.85\linewidth]{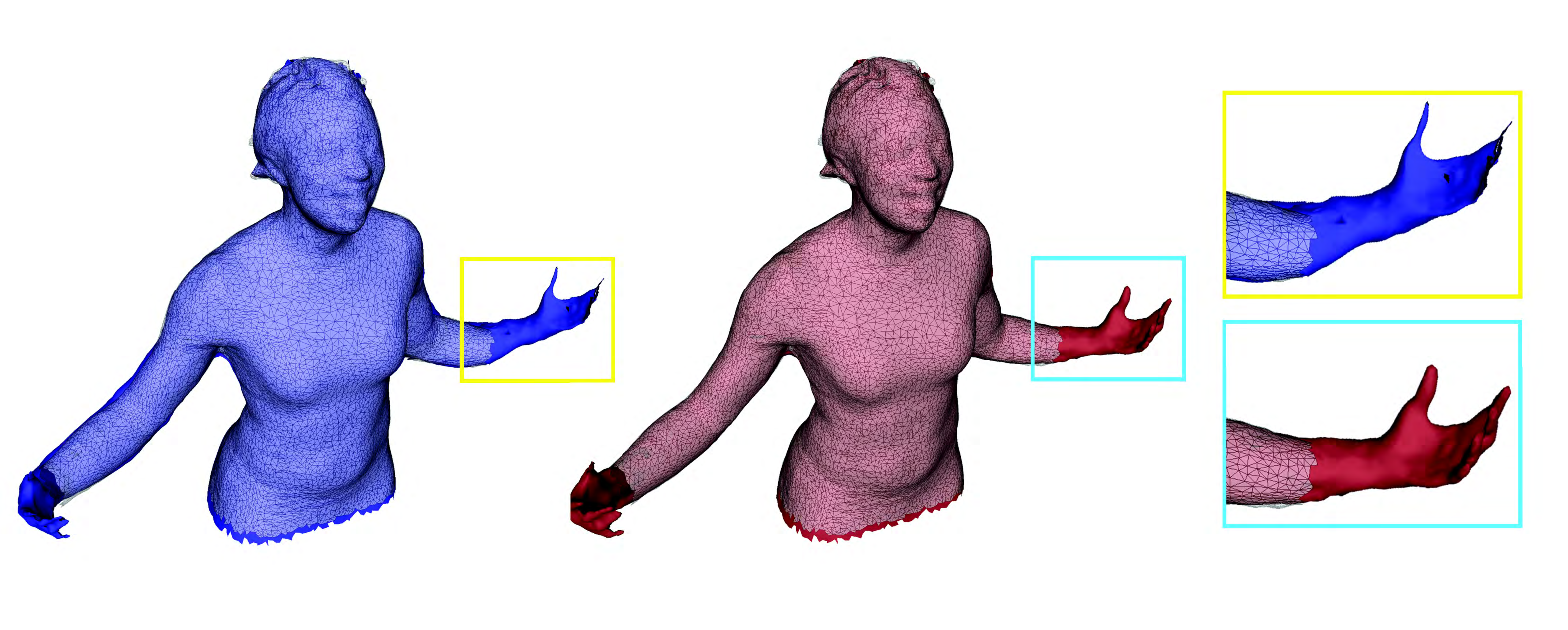}
\caption{Our warping result (right) in comparison with direct non-rigid registration result (left).}
\label{fig:warp_result_compare}
\end{figure}
After we warp the expanded global template $\boldsymbol{P}_{expan}^n$ back to each input frame $\boldsymbol{P}^{n}$, we obtain the initial warped position of the global template $\small\{\boldsymbol{P}^{0,n}\small\}_{n=1}^N$. To ensure the temporal coherency between each frame, we further adjust the motion parameter $\small\{\small\{\boldsymbol{A}_t^{0,n},\boldsymbol{b}_t^{0,n}\small\}_{t=1}^{\boldsymbol{K}}\small\}_n^N$ globally from each expanded global template $\boldsymbol{P}_{expan}^n$ to each input frame $\boldsymbol{P}^n$ by introducing a temporal term $\boldsymbol{\tilde{E}}_{tempo}$ and a data term $\boldsymbol{\tilde{E}}_{data}$ to improve smoothness:
			\begin{equation}
			\begin{split}
			 \boldsymbol{\tilde{E}}_{tempo} =  \sum_{n=2}^{N-1}||f(\boldsymbol{P}_{expan}^{n+1},\boldsymbol{A}^{0,n+1},\boldsymbol{b}^{0 \rightarrow n+1}) \\
             					+ f(\boldsymbol{P}_{expan}^{n-1},\boldsymbol{A}^{0 \rightarrow n-1},\boldsymbol{b}^{0 \rightarrow n-1})\\
                                -2f(\boldsymbol{P}_{expan}^{n},\boldsymbol{A}^{0 \rightarrow n},\boldsymbol{b}^{0 \rightarrow n})||^2
             \end{split}
             \end{equation}
             \begin{equation}
			\begin{split}
			 \boldsymbol{\tilde{E}}_{data} =  \sum_{n=2}^{N-1}||f(\boldsymbol{P}_{expan}^{n},\boldsymbol{A}^{0 \rightarrow n},\boldsymbol{b}^{0 \rightarrow n})
             				- \boldsymbol{P}_{expan}^{n}||^2
             \end{split}
             \end{equation}
where $\boldsymbol{\tilde{E}}_{data}$ forces adjust vertices to be close to the initial approximation.

Finally, we construct the energy function as $\boldsymbol{\tilde{E}}_{ajust} = \alpha_{r}\boldsymbol{\tilde{E}}_{rigid} + \alpha_{s}\boldsymbol{\tilde{E}}_{smooth} +\alpha_{t}\boldsymbol{\tilde{E}}_{tempo}+\alpha_{d}\boldsymbol{\tilde{E}}_{data}$. In our experiment, $\alpha_{r} = 100$, $\alpha_{s} = 30$, $\alpha_{t} = 1$ and $\alpha_{d} = 5$ and solve for the optimal results via Gauss-Newton technique.


Fig. ~\ref{fig:warp_result_compare} shows the warping back result use our weighted node approximation. And we also use non-rigid registration as the comparison which directly finds the warp back motion parameters from the global template to each frame. The non-rigid registration works well in overlapped regions. However, it causes severe critical bending effect in non-overlapped area. Our weighted node approximation can estimate the warping parameters accurately by combining the remaining nodes not in missing part weighted by their geodesic distance and we further use the temporal coherence constraint for further ensure the motion smoothness between adjacent frames.

\section{Experiment}
\label{sec:4}

We perform experiments on captured real-life human motion sequences to validate the effectiveness of our algorithm.

To capture high fidelity motion, we build a multi-camera system for data acquisition. Our system equipped with 20 Point Grey cameras. Each camera has resolution $1280\times720$. We have captured seven motion sequences to test our algorithm. Five sequences (Yoga, Dance, Ballet, Ballet2 and Guitar) contain full human body, eg.  and the other two sequences (Boxing and Singing) only contain the upper body. Detailed information of our test data is shown in Table.~\ref{table:data info}. All input sequences are suffered from heavy occlusion and truncation. 

\begin{table}

\centering
\caption{Data Information}
\resizebox{.8\linewidth}{!}{
\begin{tabular}{lccccccc} \toprule
    {Data}  &{\ Avg vertices} &{\ Frames} &{ Avg nodes}\\ \midrule
    Dance   & 32K & 160 & 353  \\
    Yoga    & 31K & 361 & 340   \\
    Ballet  & 29K & 200 & 350   \\
    Ballet2 & 27K & 230 & 332   \\
    Boxing  & 23K & 20 &  323    \\
   Singing  & 19K & 23 &  342   \\
    Guitar  & 29K & 21 &  356   \\
    \bottomrule
\end{tabular}
}
\label{table:data info}

\end{table}

\begin{figure}
\centering
\includegraphics[width=.85\linewidth]{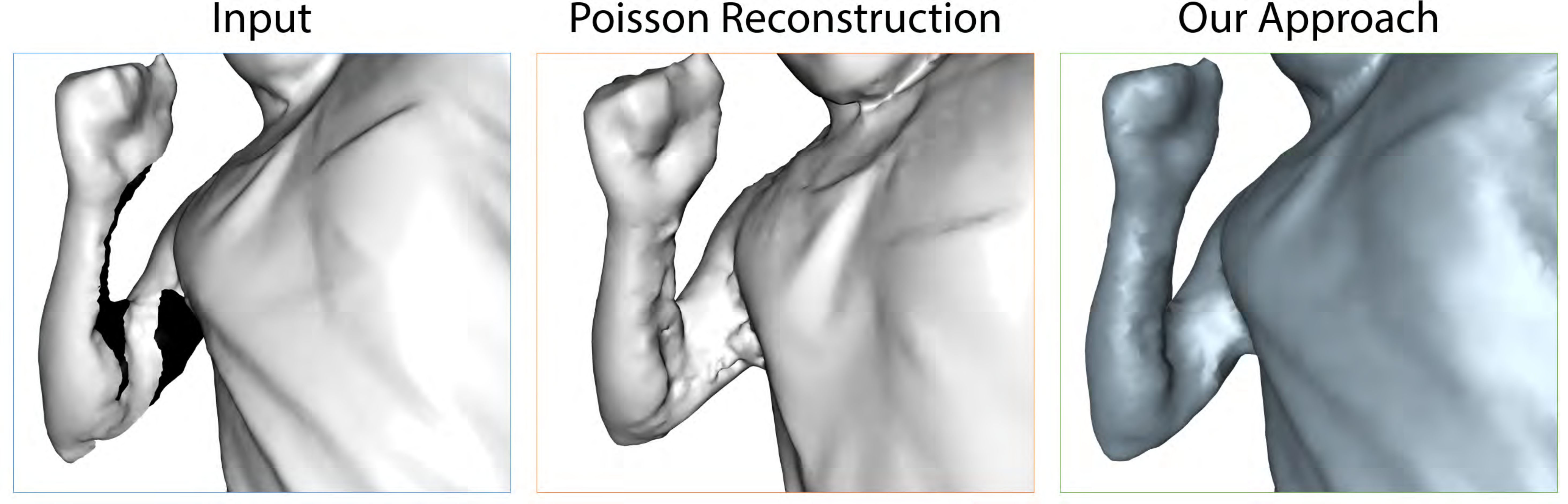}
\caption{Our reconstruction in comparison with the poisson reconstruction.}
\label{fig:compare_poisson}
\end{figure}

In pre-processing steps, we first recover a sparse point cloud using the Patch-Based Multi-View Stereo (PMVS)~\cite{furukawa2010accurate}. We then use Poisson surface reconstruction~\cite{kazhdan2006poisson} with the surface trimmer to generate an initial surface mesh. 
Due to the limited camera field-of-view and/or occlusions, the initial surface mesh might be truncated or have large holes. By taking these incomplete initial surface meshes as input, our algorithm restores the complete surface shape for every motion frame and hence recover the motion sequence with high-fidelity. We performed reconstruction using the four-step algorithm described in section \ref{sec:3}. All computations are performed off-line on a PC  with CPU Intel Core i7-5820K and 32 GB memory. In average, the running time (per frame) of our algorithm is as follow: pairwise surface matching takes around 20 seconds, \textcolor[rgb]{0.00,0.00,0.00}{global template alignment and patch segmentation costs 65 and 30 seconds respectively(both only perform once for the entire sequence), and template warping takes 10 seconds.}

We also compare our algorithm with the Poisson surface reconstruction for hole completion. The results are shown in Fig.~\ref{fig:compare_poisson}. Due to large chunk of missing data, the poisson reconstruction cannot complete the hole (\eg, the arm regions) correctly. By utilizing a global template that contains the full body geometry, our algorithm generates accurate and smooth reconstruction.


\subsection{Global Template Reconstruction Results}
\begin{figure}
\centering
\includegraphics[width=.9\linewidth]{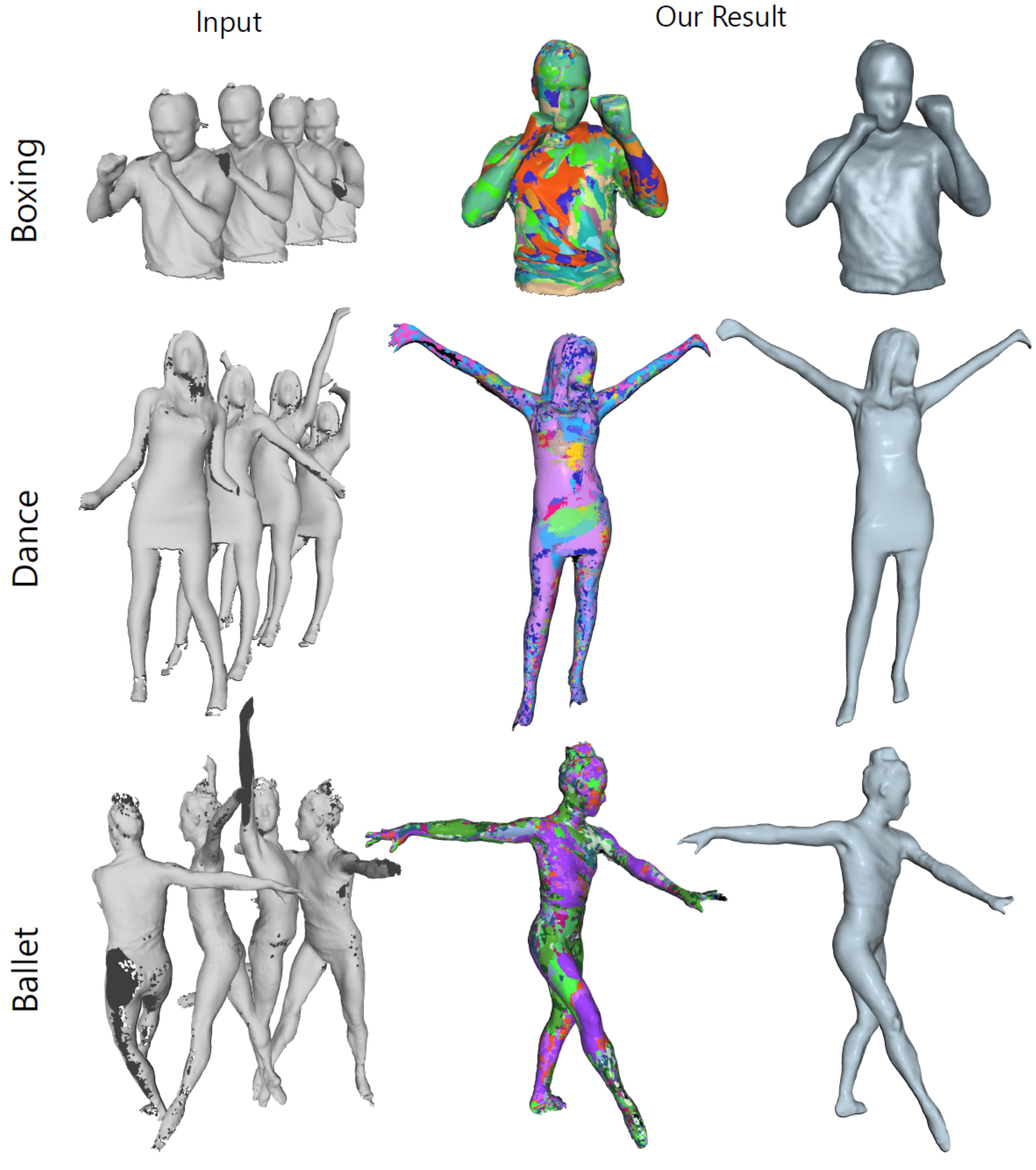}
\caption{Global template reconstruction. The first column shows corrupted input meshes from several frames. The second column shows the alignment result of the entire sequence (every frame is coded by a different color). The third column shows the global template generate by our algorithm.}
\label{fig:Template_result}
\end{figure}

In the first step, we register incomplete surfaces from the entire input sequence to generate a complete full body global template.To generate the global template, we first initiate a deformable graph for the body surface mesh of every input motion frame. We then find pairwise correspondences by imposing our temporal coherence constraint. Finally, we compute the affine transformations for every input surface mesh to align different poses and generate the global template. Fig.\ref{fig:Template_result} illustrates the global template generated by our algorithm for three different input sequences (\ie, Boxing, Dance, and Ballet). The results demonstrate that our global templates are smooth and preserve some fine details at the same time. In the second column of Fig.~\ref{fig:Template_result}, we show the composition of our global template by color-coding each frame. It shows that the poses in a motion sequences are complementary in geometry and by combining them, we are able to obtain the complete shape geometry.\textcolor[rgb]{0.00,0.00,0.00}{
When sequence is too long, Eq.\ref{eq:global} may hard to converge.So in our experiments, we set maximum frame number under 370.}

\begin{figure}
\centering
\includegraphics[width=.9\linewidth]{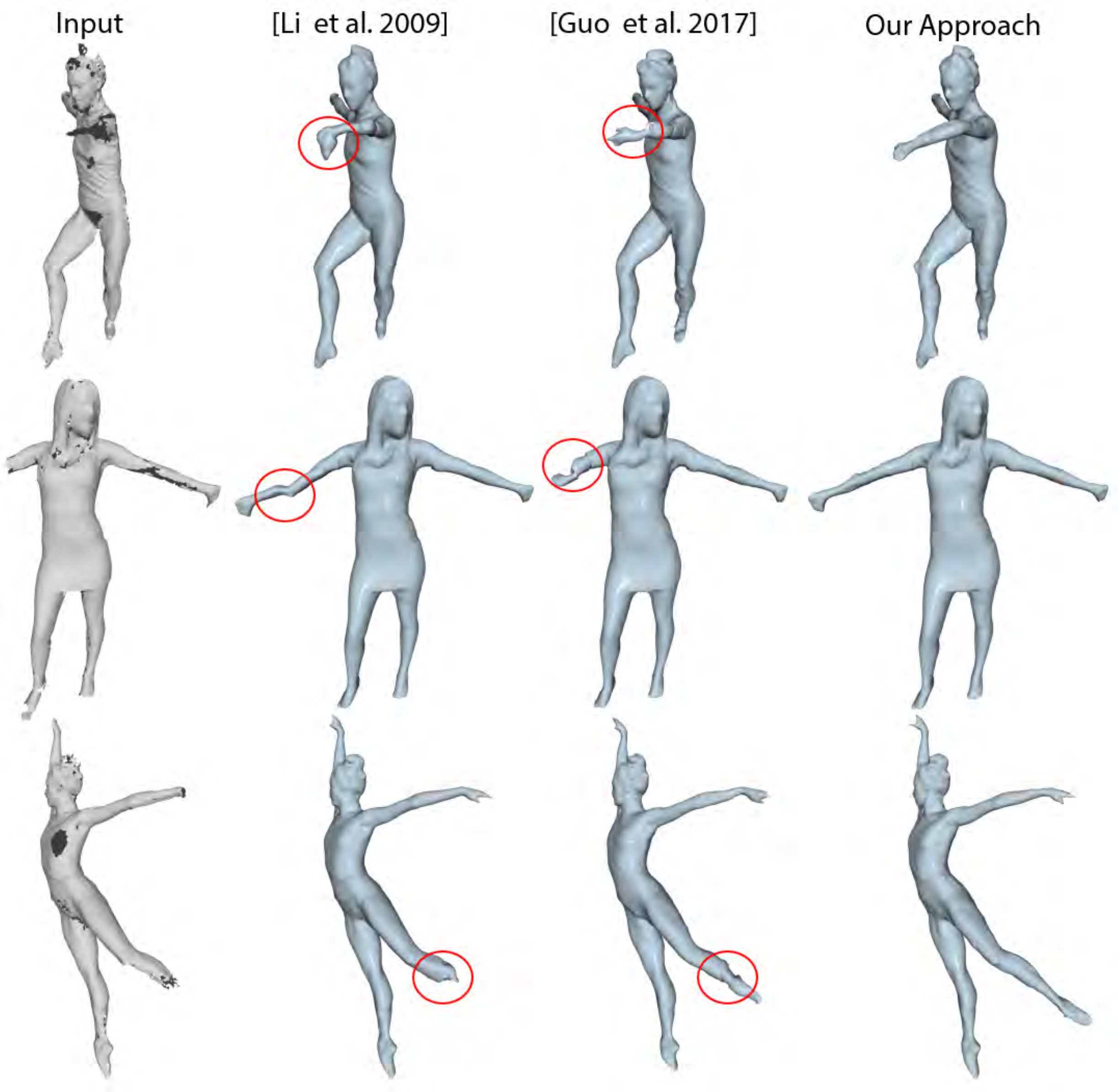}
\caption{\textcolor[rgb]{0.00,0.00,0.00}{Compare our reconstruction result with ~\cite{li2009robust} in various of scene.}}
\label{fig:result_comparison}
\end{figure}

\begin{figure}
\centering
\includegraphics[width=\linewidth]{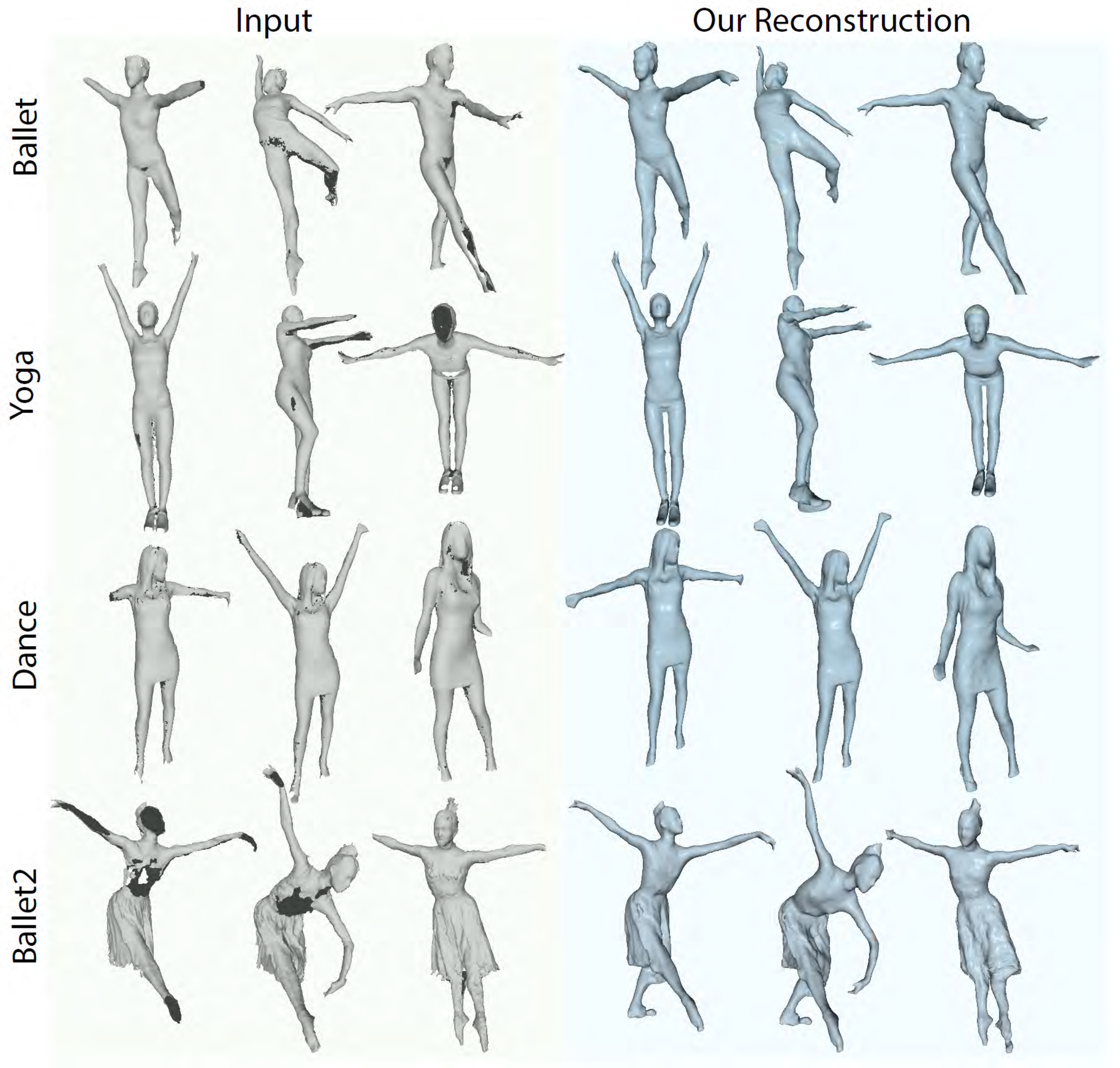}
\caption{Our motion reconstruction results. Our approach is able to restore truncated areas and fill in large holes. Please refer to the supplemental material for more results.}
\label{fig:WarpingResult}
\end{figure}

\subsection{Motion Reconstruction Results}
\begin{figure}
\centering
\includegraphics[width=.9\linewidth]{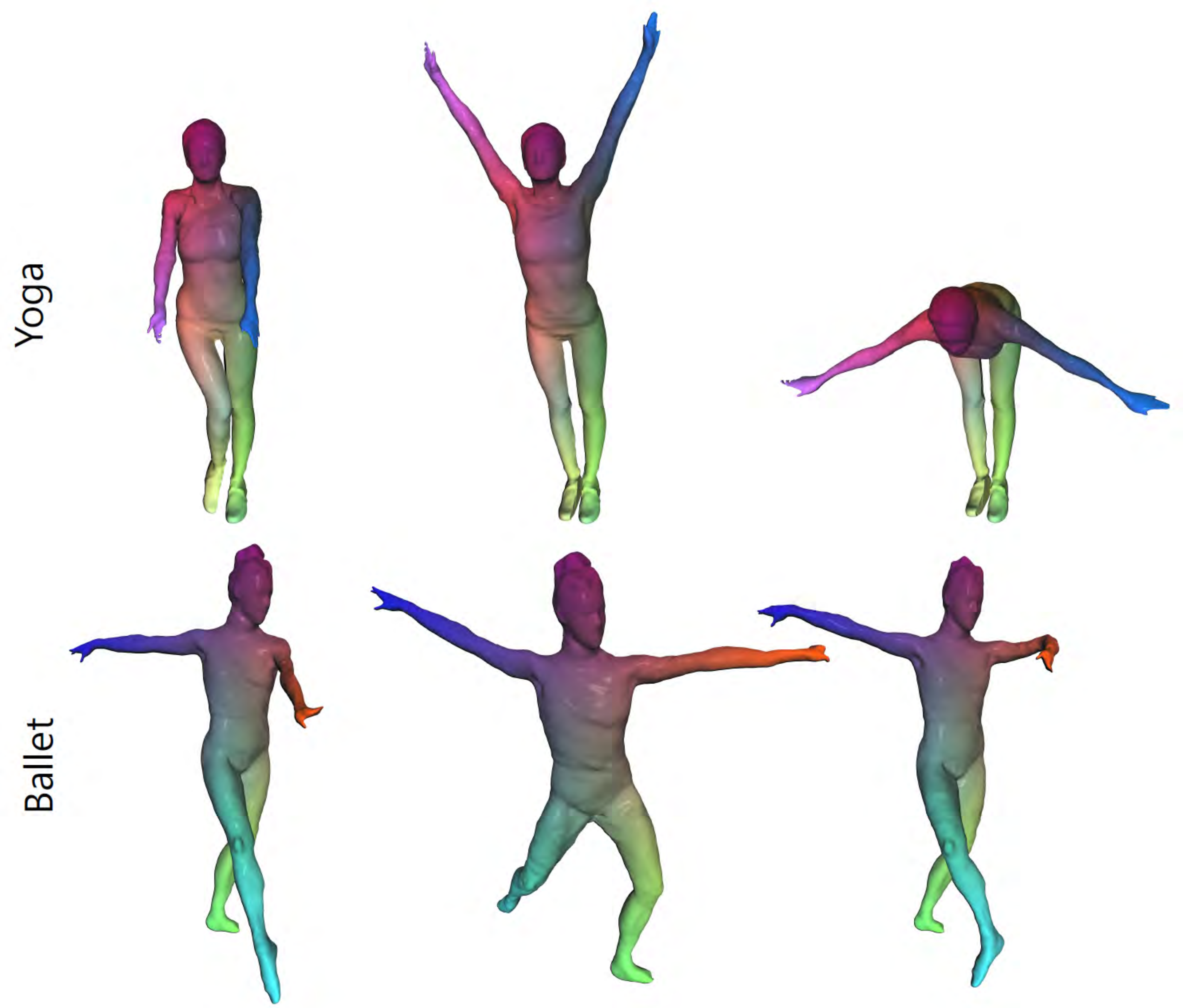}
\caption{The recovered geometry is consistent throughout the entire sequences. Corresponding vertices are color-coded.}
\label{fig:color_coded_result}
\end{figure}

\begin{figure}
\centering
\includegraphics[width=\linewidth]{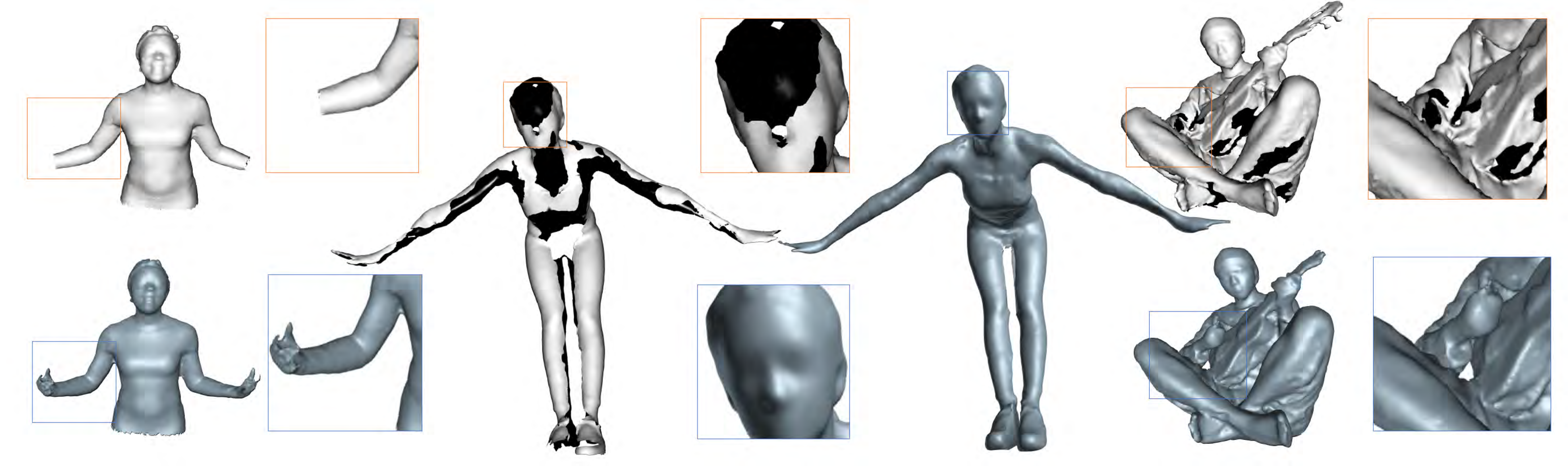}
\caption{Warping back results of singing, yoga and guitar with close-up view (from left to right).}
\label{fig:closeUpViewResult}
\end{figure}

Next, we segment the global template into connected rigid patches and build a deformable graph by taking the centroid of rigid patches as graph nodes. Finally, we warp the global template back to every input motion pose to restore the complete body surface meshes. Our reconstruction results are shown in Fig.~\ref{fig:WarpingResult}, Fig.~\ref{fig:color_coded_result} and Fig.~\ref{fig:closeUpViewResult}. Fig.~\ref{fig:closeUpViewResult} shows that our approach is capable of reconstruction full or partial body motion from heavily corrupted input data in various scenarios. Fig.~\ref{fig:WarpingResult} demonstrate that our algorithm could also handle fast, drastic and rotating motion. We can see that our approach can successfully restore truncations and fill in large holes (\eg, face in the Yoga scene, arms in the Dance scene, belly in the Ballet2 and legs in the Ballet scene). Further, because we warp the global template through entire sequences, our reconstruction results are consistent in geometry throughout the entire sequences as shown in Fig.~\ref{fig:color_coded_result}. Such consistency implies that our reconstruction could be beneficial for future applications such as consistent texture generation and data compression.

\begin{figure}
\centering
\includegraphics[width=\linewidth]{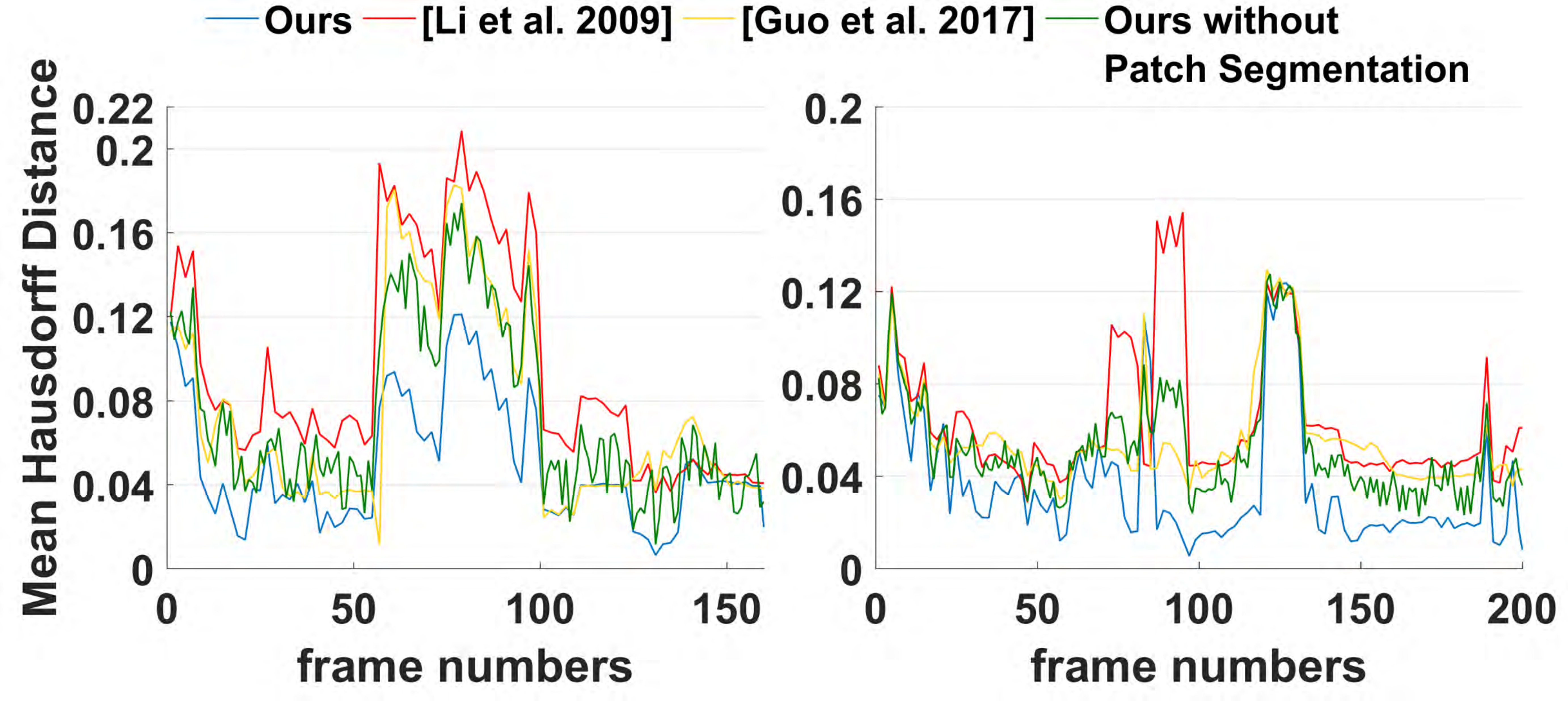}
\caption{Quantitative evaluation in comparison with \cite{li2009robust} on Ballet (a) and Dance(b).   }
\label{fig:quantitative_evaluation}
\end{figure}

 \textcolor[rgb]{0.00,0.00,0.00}{We perform experiments to compare our algorithm with the state-of-the-art method ~\cite{li2009robust} and ~\cite{guo2017robust}.} Fig.~\ref{fig:result_comparison} shows the reconstruction comparison result. We can see that our algorithm provide more accurate reconstruction in presence of large holes/truncations caused by \textcolor[rgb]{0.00,0.00,0.00}{fast motions and occlusion}. This is mainly because we consider the temporal coherence in surface alignment. We also perform quantitative evaluation to illustrate performance. In particular, we compute the mean Hausdorff distance between \textcolor[rgb]{0.00,0.00,0.00}{each pair of visible input and reconstructed surface} and use it as the metric for quantitative evaluation. \textcolor[rgb]{0.00,0.00,0.00}{We compare the distance plot of our algorithm with ~\cite{li2009robust},~\cite{guo2017robust}, and our method without patch segmentation on two input sequences(\ie Dance and Ballet ). As shown in Fig.~\ref{fig:quantitative_evaluation}, our reconstructions have lower error for most of the frames.}

\vspace{-10px}

\section{Conclusions and Discussions}
We have presented a graph-based non-rigid shape registration framework that can simultaneously recover 3D human body geometry and estimate motion at high-fidelity.Our approach is especially effective in presence of large holes and truncated areas.We propose a temporal regularization term to get more accurate pairwise correspondence than the state-of-the-art method to generate a global body template by registering all poses in the acquired motion sequence. We also developed a new segmentation algorithm to divide the global template into locally rigid patches and built a deformable graph using the rigid patches.

Our approach has several limitations. First, our proposed global template generation algorithm cannot handle topology change \textcolor[rgb]{0.00,0.00,0.00}{such as cross arms and hands}. One possible solution could be first automatically detecting topology change and then splitting the sequence into segments with the same topology and constructing separate global template. Second, subtle details(\eg, fingers) are lost in our reconstruction since they are not effectively represented in our deformable graph. To achieve even higher-fidelity, we can recover the subtle motions in a separate pass and then add them back to our reconstruction.

\section*{Acknowledgement}
This research is partially supported by National Science Foundation under the
Grant IIS-1422477 and Army Research Office
under the grant W911NF14-1-0338.

{\small
\bibliographystyle{ieee}
\bibliography{egbib}
}

\end{document}